\documentclass[conference]{IEEEtran}
\IEEEoverridecommandlockouts
\usepackage{cite}

\usepackage{amsmath,amssymb,amsfonts,mathtools,bm}
\usepackage{amsthm}
\usepackage{algorithm}
\usepackage{algorithmic}
\usepackage{graphicx}
\usepackage{textcomp}
\usepackage{xcolor,float}
\usepackage{tabularx}
\usepackage{textcomp}
\usepackage{diagbox}
\usepackage{svg}
\usepackage{booktabs}

\allowdisplaybreaks
\renewcommand{\baselinestretch}{1}

\begin{document}
    
    \title{{Digital Twin-Assisted Efficient Reinforcement Learning for Edge Task Scheduling}}

\author{
\IEEEauthorblockN{
Xiucheng Wang\IEEEauthorrefmark{1},
Longfei Ma\IEEEauthorrefmark{1},
Haocheng Li \IEEEauthorrefmark{1}, 
Zhisheng Yin\IEEEauthorrefmark{2},
Tom. Luan\IEEEauthorrefmark{2}, and 
Nan Cheng\IEEEauthorrefmark{1}\\
}
\IEEEauthorblockA{
\IEEEauthorrefmark{1}School of Telecommunications Engineering,
Xidian University, Xi'an, China\\
\IEEEauthorrefmark{2}School of Cyber Engineering,
Xidian University, Xi'an, China\\
Email: \{xcwang\_1, lfma, lhc\}@stu.xidian.edu.cn, \{zsyin, tom.luan\}@xidian.edu.cn}, dr.nan.cheng@ieee.org}
    
    \maketitle

\IEEEdisplaynontitleabstractindextext

\IEEEpeerreviewmaketitle

\begin{abstract}
Task scheduling is a critical problem when one user offloads multiple different tasks to the edge server. When a user has multiple tasks to offload and only one task can be transmitted to server at a time, while server processes tasks according to the transmission order, the problem is NP-hard. However, it is difficult for traditional optimization methods to quickly obtain the optimal solution, while approaches based on reinforcement learning face with the challenge of excessively large action space and slow convergence. In this paper, we propose a Digital Twin (DT)-assisted RL-based task scheduling method in order to improve the performance and convergence of the RL. We use DT to simulate the results of different decisions made by the agent, so that one agent can try multiple actions at a time, or, similarly, multiple agents can interact with environment in parallel in DT. In this way, the exploration efficiency of RL can be significantly improved via DT, and thus RL can converges faster and local optimality is less likely to happen. Particularly, two algorithms are designed to made task scheduling decisions, i.e., DT-assisted asynchronous Q-learning (DTAQL) and DT-assisted exploring Q-learning (DTEQL). Simulation results show that both algorithms significantly improve the convergence speed of Q-learning by increasing the exploration efficiency.
\end{abstract}


\section{Introduction}
With the development of information technologies such as Internet technology, artificial intelligence (Al), and computer vision, users are increasingly demanding services such as image recognition or VR/AR that require strong computing capability. Since such services usually have a large task size, offloading all these tasks to the cloud will result in an overburdened communication. Therefore, offloading tasks to the server in edge is emerged as an promising solution to this issue. On the other hand, due to the deployment on edge nodes, the computing capability of edge servers is limited. As a result, edge servers usually processes only a few (very likely one) tasks at the same time to provide ensuring service quality. Meanwhile, user has multiple tasks with different delay requirements that need to be offloaded, and only one task can be transmitted at a time, the order of task offloading will directly affect the quality of service. In \cite{mao2017joint} scheduling tasks is proved a flow shop scheduling problem, and in \cite{herrbach1990preemptive} this kind of problem is proved as NP-Hard. Thus, it is difficult to optimize this problem with high accuracy and speed by using tradition optimization method.

To get optimal or near-optimal solutions quickly, RL-based on task sorting algorithms have attracted increasing attention. Monte carlo tree search is applied in the specific context of task scheduling\cite{hu2019spear}. However, there is a very large action space in task scheduling problem, e.g., making it difficult for RL to achieve excellent performance. This is because usual RL agent uses $exploration-exploitation$ to converge. When agent use $exploration$, it chooses an random action through a specific distribution to obtain the performance of different permutations, and when agent use $exploitation$, it samples the best performance action repeatedly to make the algorithm converge. Obviously, $exploration$ and $exploitation$  are contradictory, the higher exploration probability, the slower it is to converge, while low exploration probability causes agent has no knowledge of enough actions, leading to a poor converge performance. Therefore, we need a better $exploration$ and $exploitation$ that has efficient exploration and high convergence speed.

DT is the virtual realization of entities in the real world through digital form. Unlike traditional simulation models, digital twins are the dynamic panoramic mapping of physical entities. Through data interaction with the entity, it constantly updates its own information to ensure the real-time, accurate and comprehensive virtual mapping, so as to realize the clone with high approximation on the virtual platform\cite{8477101}. Furthermore, the virtual model can integrate various data from the physical world, and use expert knowledge, AI and other means to conduct comprehensive analysis, so as to better predict future states. Therefore, the virtual body in the digital twin is not only a simple copy of the information of the physical entity, but a complete clone of the state, characteristics and development trend of the physical entity. When a clone is established, various complex actions can be quickly performed on the virtual platform, and obtain the state of the system after taking these actions, so as to provide support for decision-making\cite{9174795}. Compared with common methods of operating in the real world, digital twins can significantly improve efficiency and reduce costs.

Since DT can reproduce all the properties of real physical space in digital space, it enables one agent to try different actions at a time in a digital space, and also enables multiple agents to interact with the same environment in parallel come true. Therefore, to improve RL exploration efficiency, we introduce DT to get a higher performance RL with lower time. In this paper, we use DT to enable the agent to learn performance of multiple actions at a time, thus increasing the exploiting efficiency of RL-based task scheduling. The proposed methods show better performance and faster convergence speed than traditional Q-learning based RL method. To our best knowledge, this is the first paper using DT to enhance RL performance, which not only gives a way to further optimize RL-based decision methods, but also shows a potential research direction for DT.

\section{System Model}
We consider a single-user edge computing communication system\cite{mao2017joint}, where the server is deployed near the user to provide service. At the beginning of each frame, the user generates $N$ independent tasks and the task $i$ is denoted by $t_i, i \in {1,2,...N}$. Due to the limited computing power and energy of user, all tasks need to be transmitted to the edge server for execution. In this paper, we assume that the server uses a single-core CPU with constant CPU frequency, which can only process one task at a time. After the server completes a task, it immediately transmits the result back to the user.
\subsection{Task and Communication Model}
We denote the set of tasks of each communication frame by $\Gamma=\left\{t_{1}, t_{2}, \ldots, t_{N}\right\}$. Each task is described by a ternary vector $\mathbf{t_{i}}=[d_i, \varepsilon_i, c_i]$, where $d_i$ denotes the amount of the task data, $\varepsilon_i$ is the task deadline, and $c_i$ represents the task complexity which means the amount of CPU operations required to process 1 bit of data. In particular, $\varepsilon_i$ represents the maximum delay can be tolerated from task generation to result reception, which reflects the urgency of the task. The user expects to receive all the results within the deadline, otherwise the timeout tasks will be regarded as failure.

In order to avoid useless waiting time, the server executes the tasks following the order of arrival. Assuming user can only transmit the data for one task at each time. As there is only one user in this edge computing system, we consider that the channel state, transmission power and distance are known at the service, so that the transmission rates of different tasks are the same, which is denoted by $R$.

In the case of busy CPU, the task arriving at the sever will wait in the memory queue. The memory of the server only stores the data of tasks in the current communication frame. Since the amount of tasks generated by a single user in a frame is limited, we consider the memory of the server is large enough so that there will be no data overflow. Therefore, the order of task execution is the same as the order of data transmission. The computing frequency of CPU is denoted by $f_{\text {ser }}$. In addition, since the size of result data is much smaller than the input data, we ignore the result transmission delay. For the user, the most important concern is the task completion rate and total completion delay, which depends on the queue order. We denote the queue of $N$ tasks as $\boldsymbol{\sigma}=\left[\sigma_{1}, \sigma_{2}, \cdots \sigma_{N}\right]$, where $\sigma_{i}$ is the task of order $i$, $\mathrm{\sigma_{i}} \in\{1, \cdots N\}$. Consequently, our optimization goal is to ensure that all tasks can be completed within the deadline, and to minimize the total completion delay.

\subsection{Problem Formulation}
The completion time of task $t_{j}$ is denoted by $T_{\text {comp }}^{j}(\boldsymbol{\sigma})$, which includes the delay of transmission, execution, and queuing. The execution of a task only begins when all of its data has been received by the server and there are no other tasks ahead of the queue. So only when $t_{j}$ is at the head, it will be transmitted and executed without any wait. Other tasks must wait before transmission. We denote the whole data input time of the $j$-th task by $T_{\text {ready }}^{j}(\boldsymbol{\sigma})$, which is given by
\begin{align}
    T_{\text {ready }}^{j}(\boldsymbol{\sigma})=\sum_{i = 1}^j \frac{d_{\sigma_{i}}}{R}, j=1, \cdots, N.
\end{align}
Obviously, $T_{\text {comp }}^{j}(\boldsymbol{\sigma})$ includes $T_{\text {ready }}^{j}(\boldsymbol{\sigma})$ and the delay in the server which depends on $T_{\text {ready }}^{j}(\boldsymbol{\sigma})$ and the completion time of the task ahead. Thus, it can be determined by

\begin{align}
    T_{comp }^{j}(\boldsymbol{\sigma})=\left\{\begin{array}{rr}
        T_{\text {ready }}^{j}(\boldsymbol{\sigma})+d_{\sigma_{j}} c_{\sigma_{j}} f_{\text {ser}}^{-1}, & j=1 \\
        \max \left\{T_{\text {ready }}^{j}(\boldsymbol{\sigma}), T_{\text {comp }}^{j-1}(\boldsymbol{\sigma})\right\} & \\
        +d_{\sigma_{j}} c_{\sigma_{j}} f_{\text {ser }}^{-1}, & j>1
        \end{array}\right.
\end{align}
where $d_{\sigma_{j}}$ and $c_{\sigma_{j}}$ represent the data size and complexity of $t_{\sigma_{j}}$ respectively, and when $j=N$, $T_{\text {comp }}^{j}(\boldsymbol{\sigma})$ is the total completion time of the task sequence.

In order to simultaneously optimize the task success rate and total completion time within a frame, we formulate the queuing problem as

\begin{align}
     \min_{\boldsymbol{\sigma}} \sum_{i=1}^N T_{comp}^i + \zeta \mathbb{I}(T_{comp}^i > \varepsilon_i),
\end{align}
where $\mathbb{I}(\ast)$ is indicator function, when $\ast$ is true $\mathbb{I}(\ast)=1$, otherwise $\mathbb{I}(\ast)=0$, $\zeta$ is importance factor.

\section{DT-Assisted Q-learning Method}
\subsection{Q-learning}
Reinforcement learning has become a powerful means to solve the problem of resource management and task scheduling in wireless communication networks\cite{8672604,9222519,9547838}, where Q-learning\cite{8836506} is a value-based algorithm which has good convergence. The goal of reinforcement learning is to get the optimal policy which means helping the agent maximize the reward value. In most cases, the agent needs to take a series of actions to complete a task and the reward are delayed, so the expected total reward Q is seen as the evaluation function of the action, which is given by $Q=\sum_{t=1}^{\infty} \gamma^{t-1} r_{t}\left(s_{t}, a_{t}\right)$, where $\gamma$ is the discount factor, which represents the influence of future states on the current policy, $s_{t}$ and $a_{t}$ are the state and action at time $t$, respectively, $r_{t}\left(s_{t}, a_{t}\right)$ is the present reward for taking $a_{t}$ in $s_{t}$.
Q-learning stores Q values by building a Q-Table whose coordinates are states and actions. With the table, the agent only needs to select the action with largest Q value to execute according to the state of environment. However, it is often difficult to obtain the accurate Q-Table due to the unknowns of the environment.

For large state spaces, Q-learning uses Temporal-Difference method to approximate the true Q value. Specifically, when the agent observes state $s_{t}$ at time t, the action is selected by the $\epsilon$-greedy strategy, where in the greedy decision, the optimal action is given by
\begin{align}
a_{t}^{opt}=\arg \max _{a_{i} \in A_{t}} Q\left(s_{t}, a_{i}\right),
\end{align}
where $a_{t}^{opt}$ and $A_{t}$ are the optimal action and the set of actions at time $t$, respectively. 
Execute the selected action and enter the next state, then update the Q value according to the feedback of the environment, which is formulated as
\begin{align}
Q\left(s_{t}, a_{t}\right) \leftarrow  Q\left(s_{t}, a_{t}\right) \notag
&+\alpha\left[r_{t+1}\left(s_{t}, a_{t}\right)+\gamma \max _{a} Q\left(s_{t+1}, a\right)\right], \label{update} \\
\end{align}
where $\alpha$ is the learning rate, which is used to balance the efficiency and stability of learning. Based on the above method, iterate the Q value  until get a reliable Q-Table.

A key point of Q-learning is that the agent adopts the $\epsilon$-greedy strategy to choose actions, which can balance $exploitation$ and $exploration$ well. In details, $\epsilon$-greedy means that when the agent makes a decision, there is a small probability $\epsilon(\epsilon<1)$ to randomly select an action, and choose the largest-value action with the probability of remaining $1-\epsilon$. Assume that the initial state is $s_{1}$, the set of available actions is $A_{1}$ and the known optimal action is $a_{1}^{opt}$. After the agent takes an action, it receives a reward $r_{1}$ from environment. In the decision-making process, the probability of each non-optimal action being selected is $\frac{\epsilon}{\left|A_{1}\right|}$, where $\left|A_{1}\right|$ represents the number of actions. The probability of choosing $a_{1}^{opt}$ is $\frac{\epsilon}{\left|A_{1}\right|}+1-\epsilon$. Therefore, by using the $\epsilon$-greedy strategy, the agent can make a good trade-off between $exploitation$ and $exploration$.

\begin{algorithm}[!h]
	\caption{Digital Twin-Assisted Asynchronous Q-learning Algorithm}
	\begin{algorithmic}[1]
	\STATE Initialize all elements in Q-Table as $0$, learning rate $lr$, and update date cycle $\delta$
    \FOR{$i=1$ to epoch}
    \STATE $\epsilon = \epsilon_{min} + \epsilon e^{-i\beta}$
    \STATE With probability $\epsilon$ select a random action $a_{r}$, otherwise select $a_r = \arg \max_a \mathcal{Q}(s_t,a)$
    \STATE Take action $a_r$ in real environment and get reward $r_r$ and next state $s_{t+1,r}$
    \STATE $\mathcal{Q}_r (s_t,a_r) = \mathcal{Q} (s_t,a_r) + lr(r+\gamma\max_a\mathcal{Q}_r (s_{t+1},a)-\mathcal{Q}_r (s_t,a_r) )$
    \FOR{$j=1$ to $\phi$}
    \STATE $j$-th agent select a random action $a_{j}$ with probability $\epsilon$, otherwise select $a = \arg \max_a \mathcal{Q}_j(s_t,a)$
    \STATE $\mathcal{Q}_j (s_t,a_j) = \mathcal{Q}_j (s_t,a_j) + lr(r+\gamma\max_a\mathcal{Q}_j (s_{t+1},a)-\mathcal{Q}_j (s_t,a_j) )$
    \ENDFOR
    \IF{$i\%\delta==0$}
    \STATE $\mathcal{Q} = \frac{\mathcal{Q} + \sum_{j=0}^{\phi}\mathcal{Q}_j}{1+\phi}$
    \STATE $\mathcal{Q}_r = \mathcal{Q}$
    \FOR{$j=1$ to $\phi$}
    \STATE $\mathcal{Q}_j = \mathcal{Q}$
    \ENDFOR
    \ENDIF
    
    \ENDFOR
	\end{algorithmic}
\end{algorithm}

\begin{algorithm}[!h]
	\caption{Digital Twin-Assisted Exploring Q-learning Algorithm}
	\begin{algorithmic}[1]
	\STATE Initialize $\mathcal{Q} (s,a)$ as $0$ and learning rate $lr$
    \FOR{$i=1$ to epoch}
    \STATE $\epsilon = \epsilon_{min} + \epsilon e^{-i\beta}$ 
    \STATE With probability $\epsilon$ select a random action $a_{r}$
    \STATE otherwise select $a_r = \arg \max_a \mathcal{Q}(s_t,a)$
    \STATE Take action $a_r$ in real environment and get reward $r_r$ and next state $s_{t+1,r}$
    \STATE Randomly select $\phi$ unique actions $a_1,a_1,\cdots ,a_\phi$
    \STATE Take these action respectively in DT to get reward $r_1,r_2,\cdots, r_\phi$ and next state $s_{t+1,1},s_{t+1,2},\cdots,s_{t+1,\phi}$
    \STATE
    \STATE /* Update Q-Table */
    \STATE $\mathcal{Q} (s_t,a_r) = \mathcal{Q} (s_t,a_r) + lr(r+\gamma\max_a\mathcal{Q} (s_{t+1},a)-\mathcal{Q} (s_t,a_r) )$
    \FOR{$j=1$ to $\phi$}
    \STATE $\mathcal{Q} (s_t,a_j) = \mathcal{Q} (s_t,a_j) + lr(r+\gamma\max_a\mathcal{Q} (s_{t+1},a)-\mathcal{Q} (s_t,a_j) )$
    \ENDFOR
    \ENDFOR
	\end{algorithmic}
\end{algorithm}

\subsection{DT-Assisted Method}
\begin{figure}[h]
  \centering
  \includegraphics[width=0.9\columnwidth]{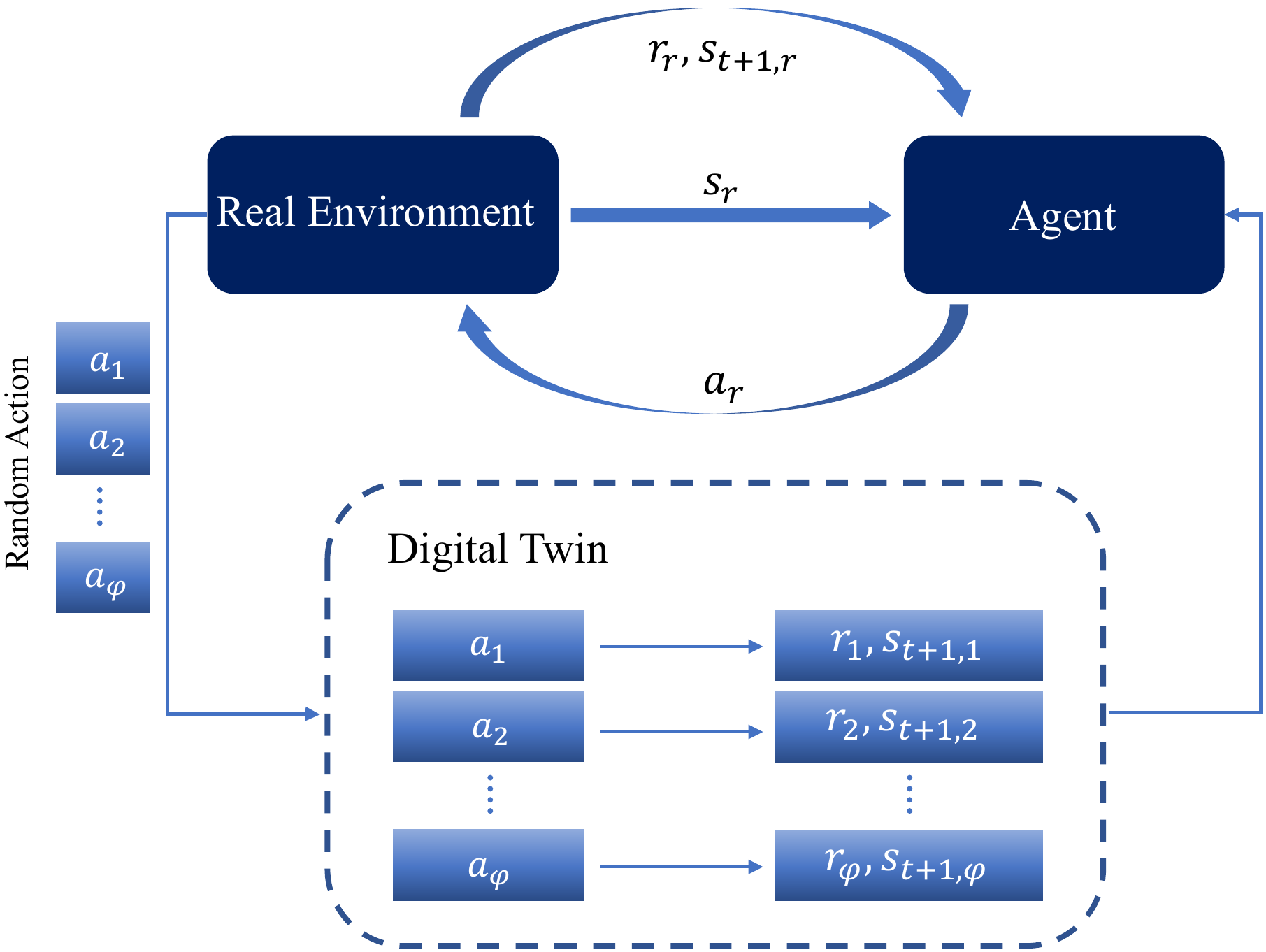}
   \vspace{-5pt}
  \caption {DT-assisted exploring Q-learning algorithm flowchart.}
   \vspace{-9pt}
\end{figure}

Traditional Q-learning can only interact with the real physical environment and can only explore one action at a time, so the exploration efficiency is low. In addition, since the real physical environment can only accept decisions made by one agent, the agent can only converge in one direction, and it is easy to fall into a local optimum. Inspired by \cite{a3c} we first tried to place multiple agents in DT, which means DT-assisted asynchronous Q-learning (DTAQL). Assuming that the simulation capability of DT is $\phi$, which means DT can complete the simulation of $\phi$ times interaction between the agent and the environment within a tolerable time. Therefore, we assume there are $\phi$ agents in DT, and each agent maintains a Q-Table, and independently selects actions in DT to interact with the environment according to the $\epsilon$-greedy algorithm. Then, each agent updates its Q-Table according to the actions it takes and the rewards the get. When all agents have updated the Q-Table $\delta$ times, they share their knowledge and update the Q-Table. Since  all agents are independent of each other, they may take different actions, which improves the efficiency of exploration. Besides, through periodically sharing knowledge, the agents in the real physical environment can obtain rewards of different scheduling order faster, thereby improving the convergence speed. Because knowledge sharing is periodic, each agent may take completely different actions in one cycle and update its own Q-Table independently. This means that in one cycle, the convergence direction of different agents may be different, which is beneficial for the agents in the real physical environment to avoid getting stuck in local optima and obtain better convergence performance.

However, asynchronous reinforcement learning needs to maintain multiple Q-Tables, which is memory consuming. Due to periodically sharing knowledge, agents are not completely independent. This leads to the fact that although the convergence directions of the agents in one cycle may be different, but on a large scale all agents still converge in the same direction. Thus, the efficiency of exploration is limited, and the convergence speed and performance is degraded. 

Therefore, we proposed another DT-assisted exploring Q-learning (DTEQL) method. In this method, only one agent is needed. First, the agent selects an action $a_r$ to act on the real environment according to the $\epsilon$-greedy algorithm, and obtains the $r_r$ and $s_{t+1,r}$ of the real environment feedback. At the same time, randomly select different $\phi$ actions $a_1, a_2,\cdots ,a_\phi$, and apply these $\phi$ actions to the virtual environment of DT respectively to obtain reward $r_1,r_2,\cdots, r_\phi$ and next state $s_{t+1,1},s_{t+1,2},\cdots,s_{t+1,\phi}$. Then, according to Equation \ref{update}, updates the Q value of these actions. Compared with DT-assisted asynchronous Q-learning, this method is more random in action selection thus improving the exploration efficiency. Moreover, because the agent always interacts with the real environment according to the $\epsilon$-greedy algorithm, the probability of selecting the optimal action in the Q-Table is increased, so that the agent can converge to the optimal value faster.

\section{Simulation Result}
In this section, we provide the simulation result to show the efficiency for DT-assisted reinforcement learning method. In simulation, task size $d$, task complexity $c$, and deadline $\varepsilon$ are all assumed to distribute uniformly, $d\sim Unif[0,2Mb]$, $c\sim Unif[0,1000]$ CPU cycles and $\varepsilon\sim Unif[1,5]s$. The CPU frequency of edge server is 10GHz. We randomly generated $1,000$ tasks to evaluate the convergence speed and performance for different algorithms. The learning rate $lr$ is 0.1. We use the following algorithm for comparison.

$\bullet$ \textbf{QL}: The traditional Q-learning method using $\epsilon$-greedy to explore the environment. There is only one agent in the real physical space, and the agent can only choose one action to interact with the environment at a time. Due to the huge state space, we set the minimum exploration rate $\epsilon_{min}$ as $0.1$ and the exploration decay factor $\beta$ as $5,000$.

$\bullet$ \textbf{DTAQL}: DT-assisted asynchronous Q-learning method. Besides real physical space, $\phi$ agent is placed in the digital space to interact with the virtual environment and periodically share knowledge, and more details are presented in Algorithm 1. But when testing performance, we only test the performance of the agent in the real environment. As for $\delta$, all agent share their knowledge every $512$ iterations.

$\bullet$ \textbf{DTEQL}: DT randomly selects p different actions and simulates the outcomes of these actions, thereby increasing the exploration probability, and more details are presented in Algorithm 2. Similar to DTAQL, we only test the performance of the agent in the real environment.

\begin{figure}[h]
  \centering
  \includegraphics[width=0.9\columnwidth]{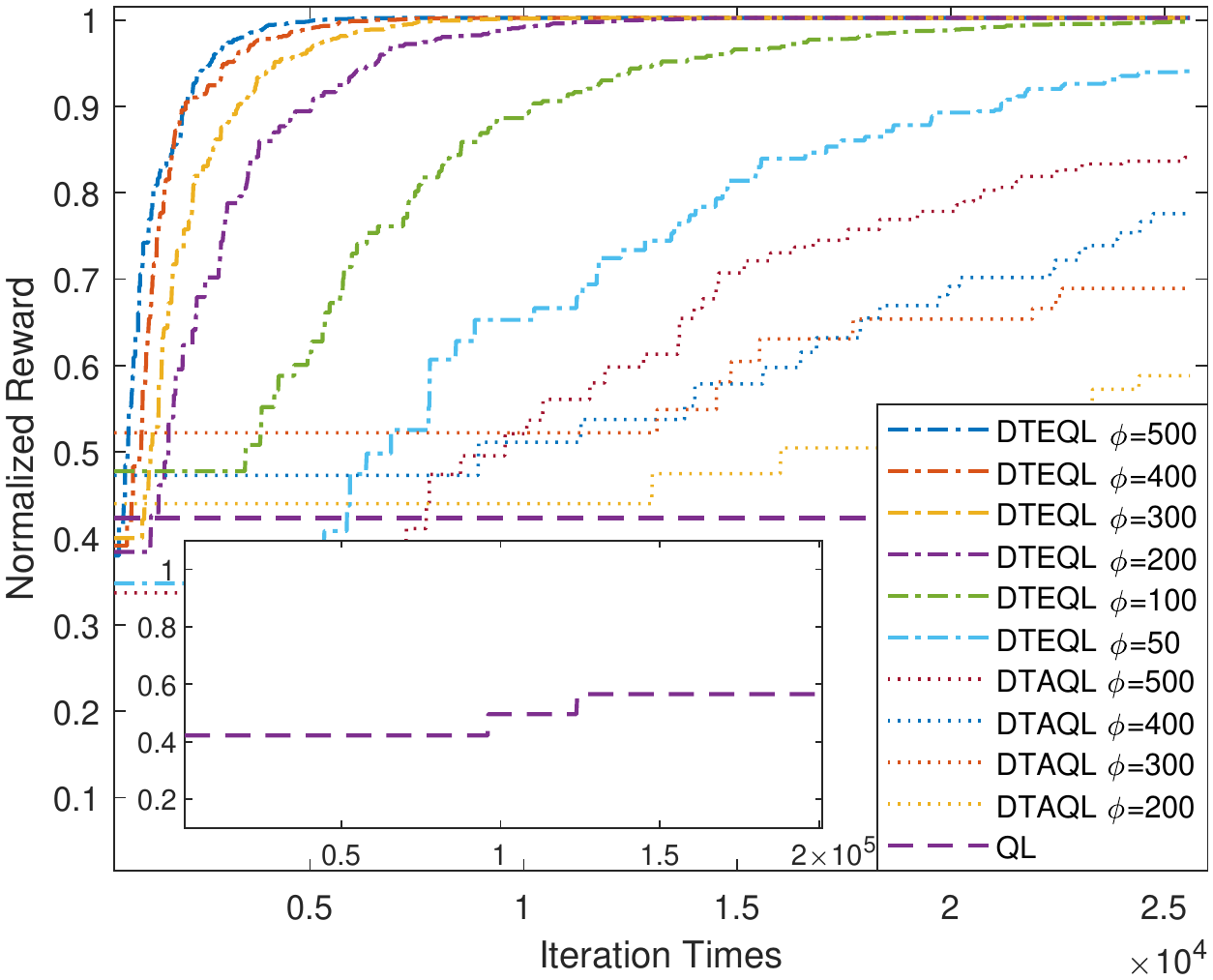}
   \vspace{-5pt}
  \caption {Convergence performance of different algorithm with $N=8$. Since QL need over 100,000 iterations to converge, while other algorithms need far much fewer iterations, so we show it in a separate figure specially.}
  \label{Reward}
   \vspace{-9pt}
\end{figure}

\begin{table}[h]
	\renewcommand{\arraystretch}{1.5}
	\newcolumntype{P}[1]{>{\centering\arraybackslash}p{#1}}
\caption{Performance of the different algorithm}
\label{tab1}
\resizebox{\columnwidth}{!}{
    \begin{tabular}{c|cccccccc}
    \toprule[0.85pt]\toprule[0.85pt]
                    & DTEQL $\phi=$500 & DTEQL  $\phi=$200 & DTEQL $\phi=$100 & DTEQL $\phi=$50 & DTAQL $\phi=$500 & QL     \\ \hline
Normalized Reward   & 1.0024   & 1.0024   & 0.9979   & 0.9406  & 0.7758      & 0.4235 \\ \hline
Deadline Miss Ratio & 0.0195   & 0.0195   & 0.0195   & 0.0273  & 0.0429    & 0.039  \\ \hline
Average Delay       & 0.6435   & 0.6435   & 0.6453   & 0.6677  & 0.7321    & 0.8697 \\ \hline
Convergence Time    & 5597     & 12752    & 25549    & $\textgreater$25600   & $\textgreater$25600     & $\gg$25600  \\ \bottomrule[0.85pt]\bottomrule[0.85pt]
\end{tabular}}
\end{table}

\begin{table}[h]
	\renewcommand{\arraystretch}{1.5}
	\newcolumntype{P}[1]{>{\centering\arraybackslash}p{#1}}
\caption{Time to converge for the different algorithms}
\label{tab2}
\resizebox{\columnwidth}{!}{
    \begin{tabular}{c|cccccccc}
    \toprule[0.85pt]\toprule[0.85pt]
                    & DTEQL $\phi=$500 & DTEQL  $\phi=$200 & DTEQL $\phi=$100 & DTEQL $\phi=$50 & DTAQL $\phi=$500 & QL     \\ \hline

$N=6$  & 102   & 295   & 510   & 1047  & 1284    & $\textgreater$25600   \\ \hline
$N=7$  & 755   & 2497   & 3873   & 8795  & 10239    & $\gg$25600  \\ \hline
$N=8$  & 5597     & 12752    & 25549    & $\textgreater$25600   & $\textgreater$25600     & $\gg$25600  \\ \bottomrule[0.85pt]\bottomrule[0.85pt]
\end{tabular}}
\end{table}

 First, we evaluate the performance of DTAQL when $N=8$, under different number of agents $\phi$ in the DT. Obvious, as is shown in Fig~\ref{Reward}, the larger $\phi$ is, the better the performance and fast convergence speed can be achieved. This is because different agents will select actions in parallel when exploring the environment\cite{a3c}. As the number of agents $\phi$ in DT increases, the possibility of agents exploring more actions will also increase, so that through periodic knowledge sharing improves the decision-making ability of the agent in the real physical environment. However, due to the periodic sharing of knowledge between agents, each agent actually has a similar exploration direction on a large scale, so the increase of agents does not significantly improve performance. Moreover, because knowledge sharing is periodic, the performance of agents in real physical environments shows a step-up trend.

Then, we test the performance of DTEQL. Similar to DTAQL, the convergence speed increases as $\phi$ increases. But when $\phi$ is with the high value region, as  $\phi$ increases, the rate at which the convergence speed improves gradually decreases. As is shown in Fig~\ref{Reward}, DTEQL converges faster for DTAQL even when phi is small than DTAQL with $\phi=500$. The reason is that DTEQL randomly select $\phi$ different actions, thereby improving the exploration efficiency, and since the agent in the real physical environment still likes the ordinary Q-learning method, using the $\epsilon$-greedy to select optimal action, which increases the sampling probability of the optimal action and improves the convergence speed. Besides, because DTEQL does not need to store multiple Q tables, its storage consumption is also less than DTAQL. 

In Table~\ref{tab1}, we show specific data on reward, average task complete delay, deadline miss ratio, and number of times required to converge for different algorithms, if no improvement in algorithm performance can be observed during training, we use $\gg 25600$ to denote it. Limited by training time, we only trained $25,600$ times for each algorithm, which is also a very long time. For DTEQL, when $\phi$ is greater than $200$, the algorithm can always converge within $25,600$ training times, and the performance after convergence is exactly the same. But as p increases, the convergence speed also increases. When $\phi$ is $100$, we can see that the reward has decreased, but deadline miss ratio keeps constant.This shows that although the task processing delay increases when DTEQL fails to converge completely, the algorithm can still ensure that all tasks are completed within the deadline. Although the performance of DTAQL is obviously better than that of the ordinary Q-learning method, it is much weaker than DTEQL in this problem.

Table~\ref{tab2} shows convergence time for different algorithm under different $N$. As the task number $N$ decreases, the action space of the scheduling problem decreases, and the algorithm requires smaller $\phi$ to converge. This means we can adaptively change $\phi$ of DT, so as to reduce the construction cost of DT, since stronger DT means more accurate data and more advanced building technology\cite{liu2021review}, which inevitably bring higher price.

\section{Conclusion}
In this paper, we have investigated the task scheduling for edge offloading with the assistance of DT. By using DT to enrich the action space, we have proposed two DT-assisted RL algorithms to let the agent try many actions at the same time or multiple agents independently interact with the environment and exchange their knowledge periodically. Simulation results have shown DT can significantly assist improving the exploration efficiency, thereby the convergence speed of Q-learning can be increased and its convergence performance can be improved. Besides, users can experience a higher quality of service with lower latency by our proposed scheme. For future research, we will study the performance of DT-assisted task scheduling in more complex network such as space-air-ground integrated networks, whose action space is continuation.
\section*{Acknowledgement}
This work was supported by the National Key Research and Development Program of China (2020YFB1807700), the National Natural Science Foundation of China (NSFC) under Grant No. 62071356, and the Fundamental Research Funds for the Central Universities under Grant No. JB210113.
\ifCLASSOPTIONcaptionsoff
  \newpage
\fi

\bibliography{ref}
\bibliographystyle{IEEEtran}

\end{document}